%% file: main.tex
\documentclass[10pt,twocolumn,letterpaper]{article}

\usepackage[pagenumbers]{cvpr} 

\usepackage{graphicx}
\usepackage{epstopdf}
\usepackage{amsmath}
\usepackage{amssymb}
\usepackage{booktabs}
\usepackage{adjustbox}
\usepackage{multirow}
\usepackage{subcaption} 
\usepackage{pifont}
\usepackage{appendix}
\usepackage[accsupp]{axessibility}  

\usepackage[pagebackref,breaklinks,colorlinks]{hyperref}

\usepackage[capitalize]{cleveref}
\crefname{section}{Sec.}{Secs.}
\Crefname{section}{Section}{Sections}
\Crefname{table}{Table}{Tables}
\crefname{table}{Tab.}{Tabs.}

\def\cvprPaperID{885} 
\def\confName{CVPR}
\def\confYear{2023}
\newcommand{\FracPartial}[2]{\frac{\partial #1}{\partial #2}}

\begin{document}

\title{PD-Quant: Post-Training Quantization based on\\ Prediction Difference Metric}

\author{
Jiawei Liu$^{1,\star,\diamond}$
~~~
Lin Niu$^{1,\star,\diamond}$
~~~
Zhihang Yuan$^{2,\dagger}$
~~~
Dawei Yang$^{2}$
~~~
Xinggang Wang$^{1}$
~~~
Wenyu Liu$^{1,\dagger}$\\
$^1$ School of EIC, Huazhong University of Science \& Technology ~~~
$^2$ Houmo AI\\
{\tt\small \{jiaweiliu, linniu\}@hust.edu.cn~~~hahnyuan@gmail.com~~~dawei.yang@houmo.ai }\\
{\tt\small \{xgwang, liuwy\}@hust.edu.cn}
}
\maketitle
\let\thefootnote\relax\footnotetext{$^\star$ Equal contribution.
$^\diamond$ This work was done when Jiawei Liu and Lin Niu were interns at Houmo AI. $^\dagger$ Corresponding authors.}

\begin{abstract}
Post-training quantization (PTQ) is a neural network compression technique that converts a full-precision model into a quantized model using lower-precision data types.
Although it can help reduce the size and computational cost of deep neural networks, it can also introduce quantization noise and reduce prediction accuracy, especially in extremely low-bit settings.
How to determine the appropriate quantization parameters (\eg, scaling factors and rounding of weights) is the main problem facing now.
Existing methods attempt to determine these parameters by minimize the distance between features before and after quantization, but such an approach only considers local information and may not result in the most optimal quantization parameters.
We analyze this issue and propose PD-Quant, a method that addresses this limitation by considering global information. 
It determines the quantization parameters by using the information of differences between network prediction before and after quantization.
In addition, PD-Quant can alleviate the overfitting problem in PTQ caused by the small number of calibration sets by adjusting the distribution of activations.
Experiments show that PD-Quant leads to better quantization parameters and improves the prediction accuracy of quantized models, especially in low-bit settings. 
For example, PD-Quant pushes the accuracy of ResNet-18 up to 53.14\% and RegNetX-600MF up to 40.67\% in weight 2-bit activation 2-bit.
The code is released at \url{https://github.com/hustvl/PD-Quant}.
\end{abstract}

\input{sections/introduction}

\input{sections/background.tex}

\input{sections/methods}

\input{sections/experiment}

\input{sections/discussion}

\input{sections/conclusion}

\section*{Acknowledgements}
This work was supported by National Natural Science Foundation of China (NSFC No. 61733007).

{\small
\bibliographystyle{ieee_fullname}
\bibliography{egbib}
}
\newpage
\section*{Appendix}
\input{sections/supplementary.tex}

\end{document}

%% file: sections/introduction.tex
\section{Introduction}
\label{sec:introduction}
Various neural networks have been used in many real-world applications with high prediction accuracy. 
When deployed on resource-limited devices, networks' vast memory and computation costs become significant challenges. 
Reducing overhead while maintaining the model accuracy has received considerable attention. 
Network quantization is an effective technique that can compress the neural networks by converting the format of values from floating-point to low-bit~\cite{hansongcompressing, gholami2021survey, liang2021pruning}.
There are two types of quantization: post-training quantization (PTQ)~\cite{park2018value} and quantization-aware training (QAT)~\cite{hubara2021accurate}.
QAT requires retraining a model on the labeled training dataset, which is time-consuming and computationally expensive. 
While PTQ only requires a small number of unlabeled calibration samples to quantize the pre-trained models without retraining, which is suitable for quick deployment.
Existing PTQ methods can achieve good prediction accuracy with 8-bit or 4-bit quantization by selecting appropriate quantization parameters.~\cite{whitepaper,krishnamoorthi,leequantization}.
Local metrics (such as MSE~\cite{choukroun2019low} or cosine distance~\cite{wu2020easyquant} of the activation before and after quantization in layers) are commonly used to search for quantization scaling factors.
These factors are chosen layer by layer by minimizing the local metric with a small number of calibration samples.
In this paper, we observe that there is a gap between the selected scaling factors and the optimal scaling factors~\footnote{We define the optimal quantization scaling factors as the factors that make the quantized model have the lowest task loss (cross-entropy loss calculated by real label) on the validation set.}.

Since the noise from quantization will be more severe at low-bit, the prediction accuracy of the quantized model significantly decreases at 2-bit. 
Recently, some methods~\cite{brecq,qdrop,AQuant} have added a new class of quantization parameters,  weight rounding value, to adjust the rounding of weights.
They optimize both quantization scaling factors and rounding values by reconstructing features layer-wisely or block-wisely.
Besides, the quantized model by PTQ reconstruction is more likely to be overfitting to the calibration samples because adjusting the rounding of weights will significantly increase the PTQ's degree of freedom.

We propose an effective PTQ method, PD-Quant, to address the above-mentioned issues.
In this paper, we focus on improving the performance of PTQ on extremely low bit-width.
PD-Quant uses the metric that considers the global information from the prediction difference~\footnote{The prediction is the processed output of the last layer, such as the probability after softmax in the classification model.} between the quantized model and the full-precision (FP) model. 
We show that the quantization parameters optimized by prediction difference are more accurate in modeling the quantization noise.
Besides, PD-Quant adjusts the activations for calibration in PTQ to mitigate the overfitting problem.
The distribution of the activations is adjusted to meet the mean and variance saved in batch normalization layers.
Experiments show that PD-Quant leads to better quantization parameters and improves the prediction accuracy of quantized models, especially in low-bit settings. 
Our PD-Quant achieves state-of-the-art performance in PTQ.
For example, PD-Quant pushes the accuracy of weight 2-bit activation 2-bit ResNet-18 up to 53.14\% and RegNetX-600MF up to 40.67\%. 
Our contributions are summarized as follows:
\begin{enumerate}
    \item We analyze the influence of different metrics and indicate that the widely used local metric can be improved further.
    \item We propose to use the information of the prediction difference in PTQ, which improves the performance of the quantized model.
    \item We propose Distribution Correction (DC) to adjust the activation distribution to approximate the mean and variance stored in the batch normalization layer, which mitigates the overfitting problem. 
\end{enumerate}

%% file: sections/background.tex
\section{Related Work and Background}
Many excellent works have been proposed to resolve neural networks' enormous memory footprint and inference latency, including knowledge distillation~\cite{gou2021knowledge,wang2021knowledge,hinton2015distilling}, model pruning~\cite{zhu2017prune,liu2018rethinking}, and model quantization~\cite{yang2019quantization,cai2017deep,polino2018model}.

We focus on model quantization, which is an effective technique for compressing neural networks.
Quantized models keep their weights and activations in low-bit data types to reduce memory and computation requirements.
We can map a real-valued tensor $x$ (weights or activations) to the integer grid according to the following equation:
\begin{align}
    \label{eq:base_quant}
    \tilde{x}=clamp\left(\lfloor \frac{x}{S} \rceil+Z; \ q_{min}, q_{max}\right),
\end{align}
\vspace{-0.5cm}
\begin{align}
    \label{eq:quant_para}
    S = \left({x_{max}-x_{min}}\right)/\left({2^b-1}\right),
\end{align}
where $\lfloor \cdot \rceil$ is the round-to-nearest operator. 
$S$ denotes the quantization scaling factors, which reflect the proportional relationship between FP values and integers.
Moreover, $Z$ is the offset defined as zero-point.
$x_{max}$ is the maximum in the vector, and $x_{min}$ is the minimum in the vector.
$\left[q_{min}, q_{max}\right]$ is the quantization range determined by bit-width.
We only consider uniform unsigned symmetric quantization, as it is the most widely used quantization setup.
Therefore, $q_{min}$ is equal to $0$ and $q_{max}$ is equal to $2^b-1$, where $b$ is the bit-width which determines the number of integer grids.
Nonuniform quantization~\cite{Mr.BiQ} is challenging to deploy on hardware, so we will not consider it in this work.
In general, we divide quantization methods into Quantization-aware training~\cite{wu2020easyquant,zhang2018lq, han2021improving, zhou2016dorefa, choi2018pact, zhang2018lq, jung2019learning} (QAT) and Post-training quantization~\cite{adaround, brecq, qdrop, wang2020towards, cai2020zeroq, zhao2019improving} (PTQ).

\subsection{Quantization-aware training}
QAT modifies the quantization noise during training processes with full labeled training datasets. 
STE~\cite{ste} solves the problem of producing zero gradients during backpropagation by employing a gradient estimator.
During the training process, ~\cite{jacob2018quantization} smoothed activation ranges by exponential moving averages.
LSQ~\cite{lsq} introduces trainable clipping threshold parameters to learn the min and max ranges of the quantization by STE.
Although QAT enables lower bit quantization with competitive results, it needs labeled datasets and amounts of computing resources. 

\subsection{Post-training quantization}
PTQ algorithms often determine the quantization scaling factors with limited calibration data through a simple parameter space search without training or fine-tuning~\cite{aciq,zhao2019improving,choukroun2019low,wu2020easyquant,wang2019haq, PWLQ}.
Metrics for searching scaling factors include MSE distance~\cite{choukroun2019low} and cosine distance~\cite{wu2020easyquant}.
~\cite{aciq} seeks the optimal clipping ranges by minimizing the difference between FP and quantized feature map.

Later, several methods have been proposed to optimize rounding values.
Although these PTQ methods introduce fine-tuning during quantization, they still differ from QAT.
QAT adjusts the model’s weights using whole labeled training dataset, while PTQ only optimizes the quantization parameters on some unlabeled data.
AdaRound~\cite{adaround} suggests a new rounding mechanism that assigns a continuous variable to each weight value to determine whether it should be rounded up or down, rather than using the nearest rounding.
BRECQ~\cite{brecq} optimizes activation scaling factors and proposes reconstructing quantization parameters block by block. 
QDrop~\cite{qdrop} imports the activation error into the reconstruction process and introduces drop operation.
These methods can achieve usable accuracy at low bits without inference overhead.
They reconstruct features by calculating the MSE distance between quantized and FP activations.
AQuant~\cite{li2022efficient} improves activation quantization strategy and the quantization performance, but has additional inference overhead.
NWQ~\cite{wangleveraging} was published at the same time as our method, so we do not compare with it.

Some methods~\cite{cai2020zeroq, choi2022s, zhang2021diversifying, zhong2022intraq, xu2020generative, haroush2020knowledge} can now quantize models without using real data.
GDFQ~\cite{xu2020generative} adopts a generator to synthesize calibration data.
ZeroQ~\cite{cai2020zeroq} uses statistics from batch normalization layers to synthesize calibration data from a Gaussian distribution.
Compared to using real calibration data, these methods often struggle to exhibit competitive performance. 
However, inspired by ZeroQ, we proposed DC in PD-Quant.

%% file: sections/methods.tex
\section{Methodology}
\label{sec:method}
\begin{figure}[t]
  \centering
   \includegraphics[width=0.95\linewidth]{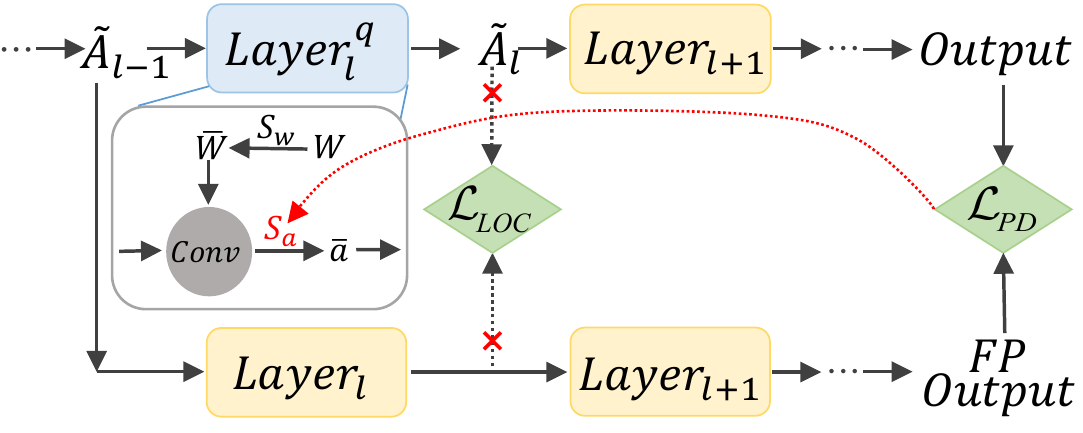}
   \caption{The pipeline overview of our method is with PD loss for determining activation scaling factors. The blue and yellow rectangle indicates the quantizing and FP layer, respectively. We mark the green diamond as the loss function. The red \ding{53} in the figure represents an approach we did not adopt but adopted in previous work.}
   \label{fig:3.1fig}
\end{figure}

In this section, we will start by conducting a comprehensive analysis of the impacts of various metrics in the search for activation quantization scaling factors.
Then, we develop a high-accuracy post-training quantization method, PD-Quant.
It applies Prediction Difference (PD) loss to optimize the quantization parameters.
Besides, we introduce regularization and propose Distribution Correction (DC) to solve the overfitting problem.

\subsection{Prediction Difference Loss}
\label{subsec:tradi_ptq}

\begin{figure}[tbp]
    \centering
    \subcaptionbox{ResNet-18\label{resnet18}}
    {
    \includegraphics[width=.48\linewidth]{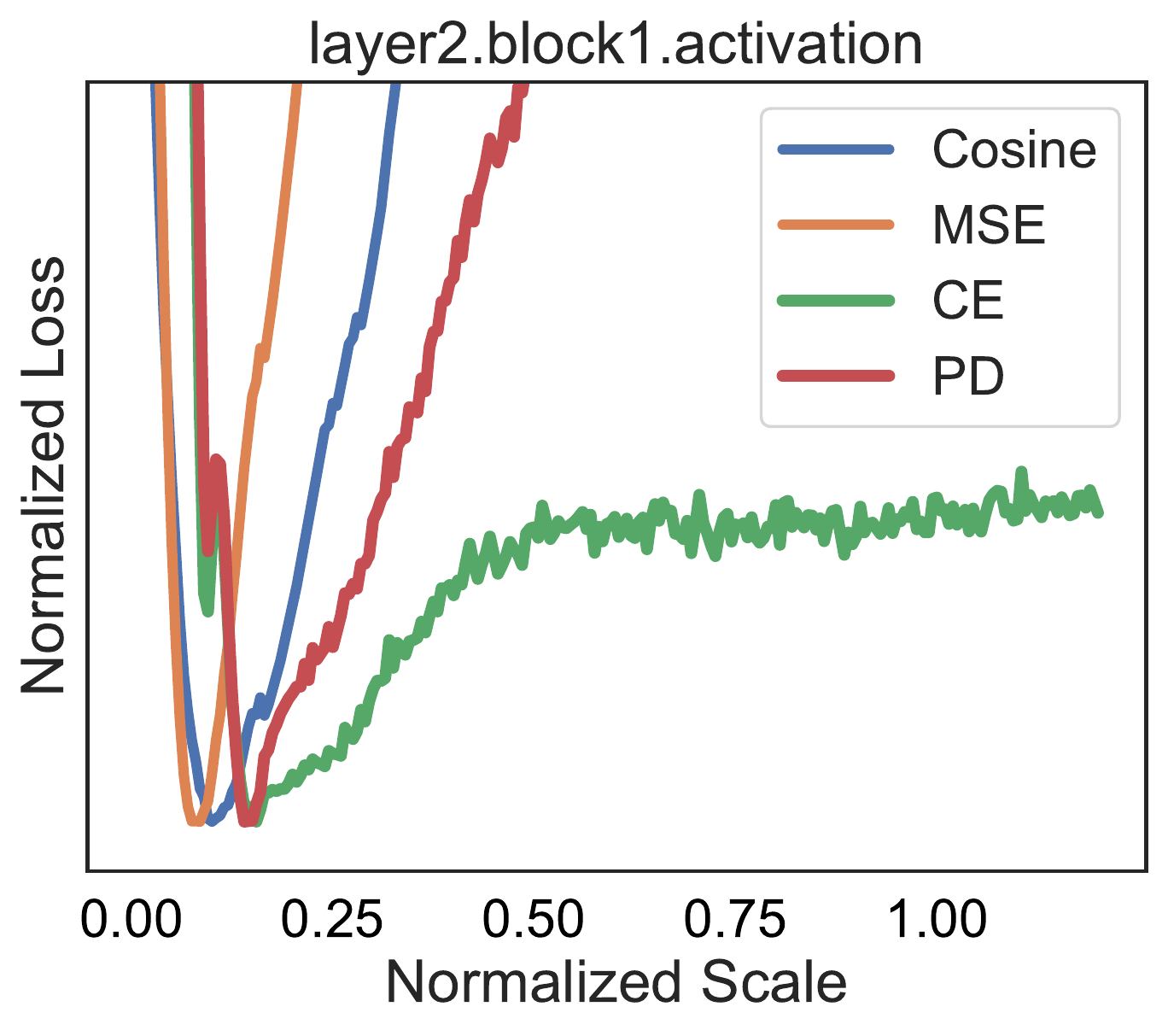}
    \includegraphics[width=.48\linewidth]{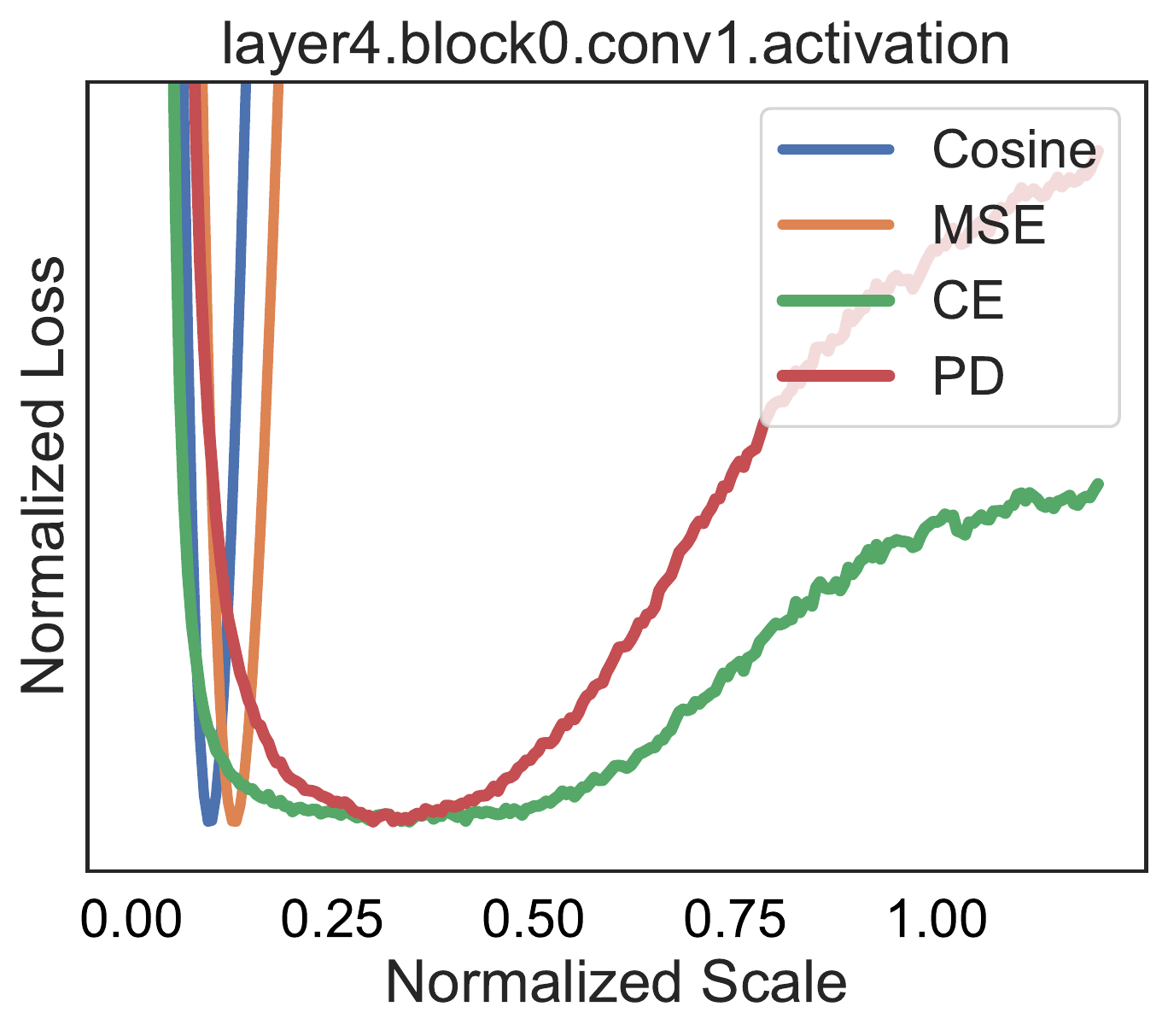}
    }\\
    \subcaptionbox{ResNet-50\label{resnet50}}
    {
    \includegraphics[width=.48\linewidth]{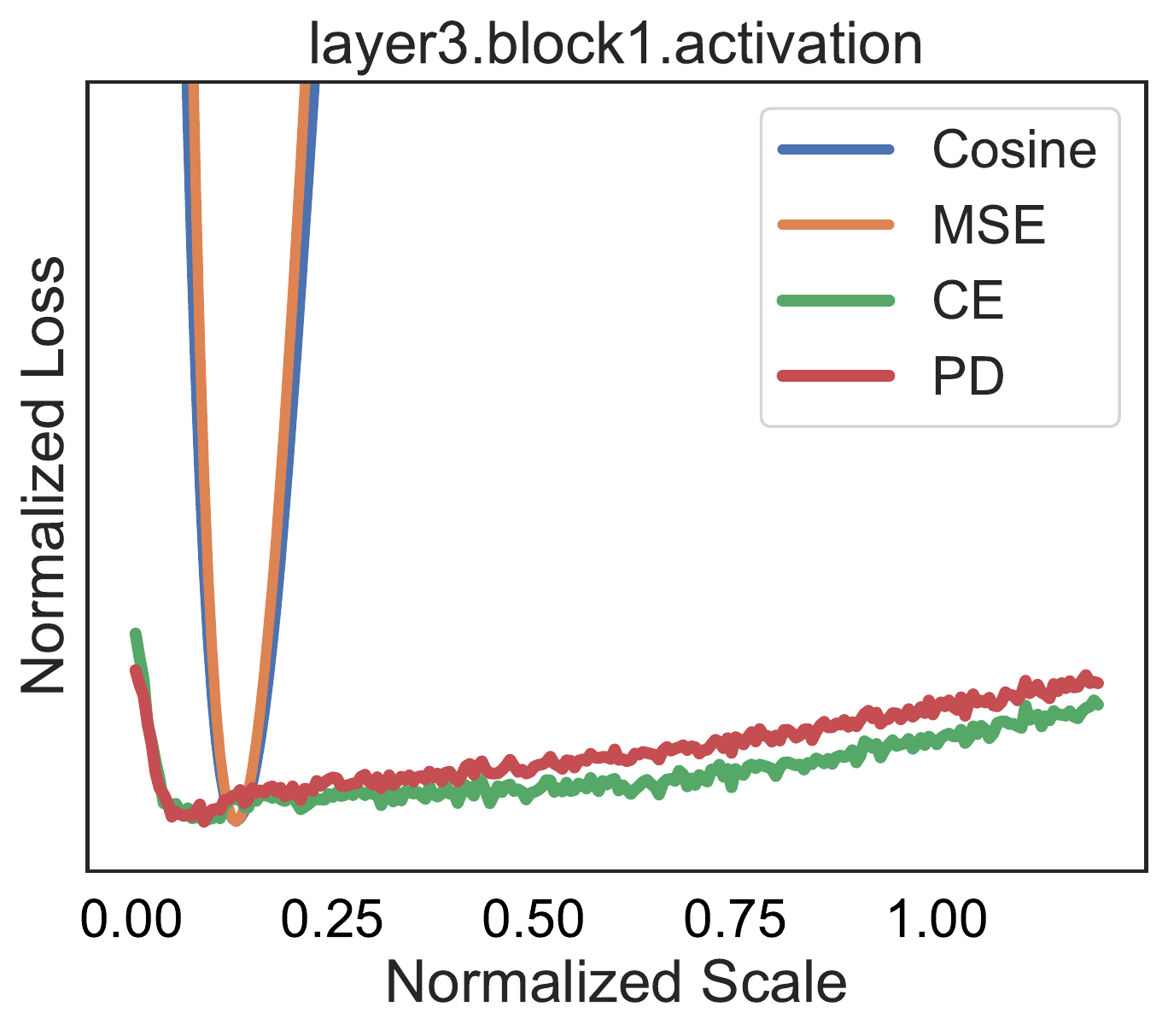}
    \includegraphics[width=.48\linewidth]{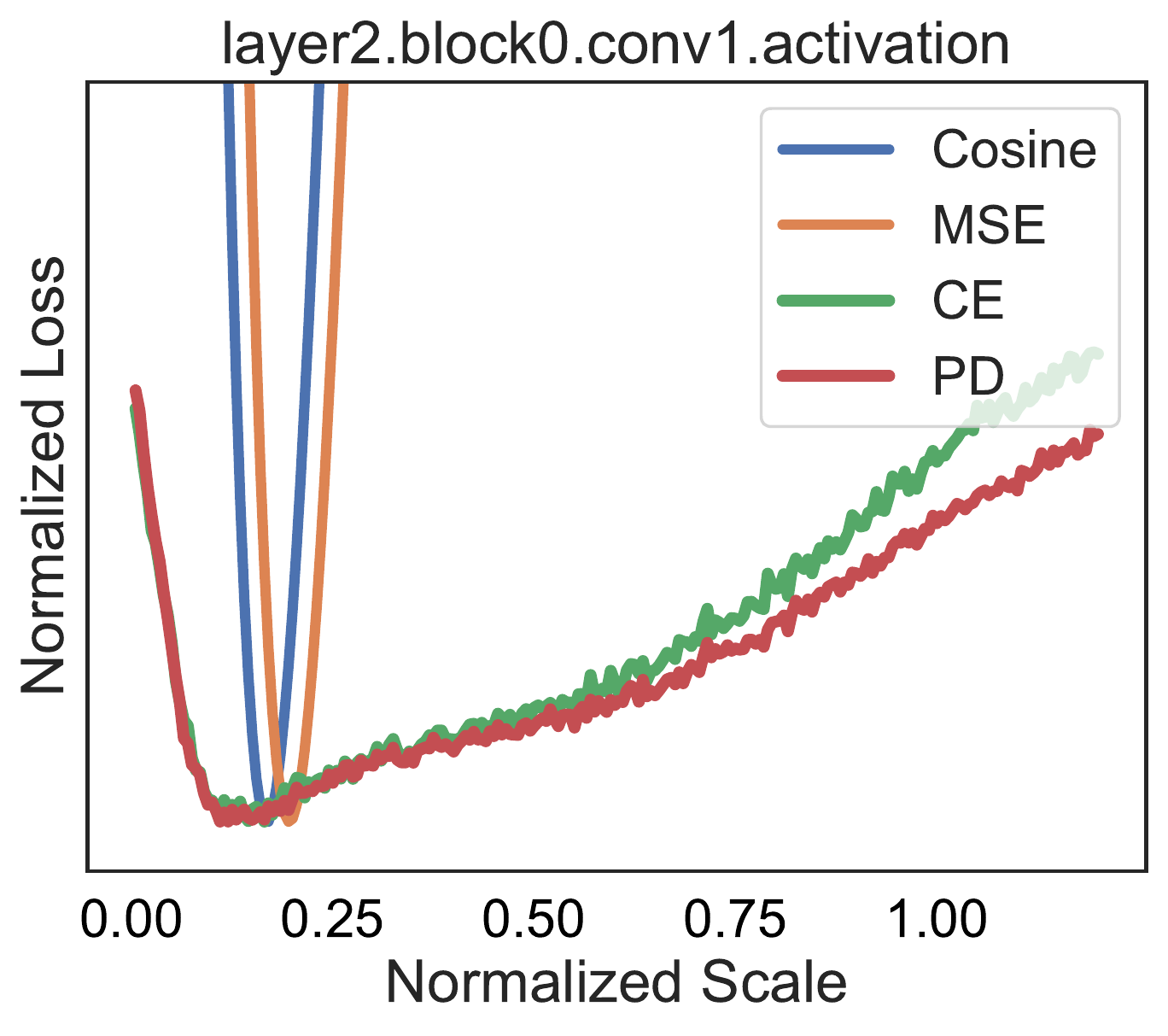}
    }
\caption{Different loss metrics are plotted, including task loss (CE) under various scaling factors on ResNet18 and ResNet50 with the ImageNet dataset. 
The normalized scaling factor $(N_s)$ means the proportion of $\frac{x_{max}-x_{min}}{2^b-1}$, and normalized loss indicates that the minimum value of all losses is normalized to 1.0.}
\label{fig:metric}
\end{figure}

Previous PTQ works~\cite{choukroun2019low,wu2020easyquant} search quantization scaling factors by local metrics, such as MSE or Cosine distance of each layer's activation before and after quantization. 
To investigate the influence of these metrics when determining activation scaling factors, we compare their search results with those of task loss.
Task loss refers to the cross-entropy loss determined by the real label, and we define the scaling factors with the lowest task loss as optimal.
\cref{fig:metric} shows the task loss and other metrics for different scaling factors of 2-bit activation. 
As seen, the scaling factor optimized by local metrics (Cosine and MSE) is inconsistent with that based on task loss. 
We can observe that the scaling factors searched by local metric losses is far from the optimal scaling factor minimized by task loss. 
As an example, as shown in \cref{resnet18}, for the activation quantization in the Resnet18.layer4.block0.conv1, green line of task loss (CE) indicates the optimal scaling factor is around $0.35 \times N_s$ (normalized scaling factor) while local metrics consider $0.15 \times N_s$ as their scaling factor.  

Since the real label is not available in PTQ, we cannot use task loss as the metric.
To achieve more accurate performance for PTQ, we propose PD loss to determine the activation scaling factors. 
PD loss compares the prediction difference between the FP and the quantized models rather than the difference in each layer's activation before and after quantization. 
Specifically, we feed the quantized activation $\tilde{A}_{l}$ into the subsequent FP layers to obtain the output prediction.
And then, the output prediction will be compared with the FP output prediction to calculate PD loss. 
\input{tables/3.1_tab}
When searching for the optimal activation scaling factors, we take the following formula as the metric:
\begin{align}\label{eq:naive_loss}
    \arg\min\limits_{S_a}\ \mathcal{L}_{PD}(f_{l}(\tilde{A}_{l-1}),\ f_{l+1}({L}^{q}_{l}(\tilde{A}_{l-1}))),
\end{align}
where $S_a$ denotes scaling factors for activation quantization in quantizing layer ${L}^{q}_{l}$, and $f_{l}(\cdot)$ refers to a part of the FP network mapping input $\tilde{A}_{l-1}$ to FP output.
$f_{l+1}(\cdot)$ is the part of the FP network mapping the output of ${L}^{q}_{l}$ to the final prediction. 
We make the assumption that the quantization for all previous layers of $\tilde{A}_{l-1}$ has been done.
\cref{fig:3.1fig} shows the detailed process for the parameters search by PD loss.

As shown in \cref{fig:metric}, the scaling factor optimized by the red line of PD loss (PD) is closer to the scaling factor expected by task loss. 
Still, take the ResNet-18 activation, as mentioned above, as an example.
The optimal scaling factor chosen by PD loss is around $0.35 \times N_s$, almost the same as CE.
In particular, we only consider the scaling factors of activation quantization, and we will discuss the scaling factors of weight quantization later in \cref{subsec:weiht_quantize}.

To further investigate how PD loss affects activation scaling factors determination, we respectively evaluate local metrics and PD loss on ResNet-18, ResNet-50~\cite{resnet}, and RegNetX-600MF~\cite{regnet} as examples.
As shown in \cref{tab:3.1_tab}, the Min-Max quantization strategy (i.e.\cref{eq:base_quant}) loses all quantization accuracy in the extremely low-bit. 
Compared to the local metric (Min-Max, MSE, Cosine), the quantization parameters optimization strategy with PD loss exhibits better results on ImageNet~\cite{russakovsky2015imagenet} dataset. 
The results verify that PD loss can significantly improve quantization performance in the extremely low-bit of the activation scaling factors.
Moreover, it indicates that the widely used local metric could be better at low-bit.
However, only optimizing the scaling factor does not gain usable results, although PD loss can achieve improvement.
So we further verified the effect of our proposed PD loss with the optimization of both activation scaling factors and rounding values in \cref{subsec:advanced_ptq}, which can achieve usable accuracy at extremely low-bit.

We choose Kullback-Leibler (KL) divergence as the PD loss metric to measure the prediction difference for the classification networks.
We will discuss the choice of PD loss in \cref{subsec:pd_determination}.
For the detailed implementation of KL, we follow~\cite{hinton2015distilling}.
In summary, according to our analysis, considering prediction differences is beneficial for searching the activation scaling factors at low-bit.

\subsection{Prediction Difference Loss with Regularization}
\label{subsec:advanced_ptq}
Recently, some PTQ methods (such as AdaRound\cite{adaround}, BRECQ~\cite{brecq}, and QDrop~\cite{qdrop}) have achieved remarkable progress on PTQ, especially in low-bit.
They propose optimizing a continuous variable for each weight value that will be quantized.
These variables, called rounding values, determine whether weight values will be rounded up or down during the quantization process.
The quantization parameters in these PTQ methods contain scaling factors and rounding values.
In summary, we can describe the process of weight quantization as follows:
\begin{align}
    \label{eq:ada_quant}
    \tilde{x}=clamp\left(\lfloor \frac{x+\theta}{S} \rceil +Z;\  q_{min}, q_{max}\right),
\end{align}
where $\theta$ is the optimization variable for each weight value to decide rounding results up or down~\cite{adaround}, \ie, $\frac{\theta}{S}$ ranges from 0 to 1. 
Other parameters are consistent with \cref{eq:base_quant}.
As for activation quantization, the quantization steps are also the same as \cref{eq:base_quant}.

The next problem is whether PD loss (as described in \cref{subsec:tradi_ptq}) can also improve the optimization of both scaling factors and rounding values.
We conduct preliminary experiments with PD loss on the ImageNet dataset with 1024 calibration samples to answer this question.
We calculate PD loss as the difference between the prediction of the FP model and the prediction of the quantized model.
However, as shown in \cref{tab:3.2_tab}, the performance of PD-only degrades severely in all settings.
Moreover, it performs much worse on the validation set than on the calibration set.
For example, the accuracy of 2 bit quantization (W2A2) on the calibration set is almost equal to the FP model on ResNet-18~\cite{resnet}, but the performance on the validation set is extremely poor.
This phenomenon indicates that the model suffers from a severe overfitting problem. 
This issue arises from the limited amount of data available for calibration.

\input{tables/3.2_tab}

\begin{figure}[t]
\centering
   \includegraphics[width=0.95\linewidth]{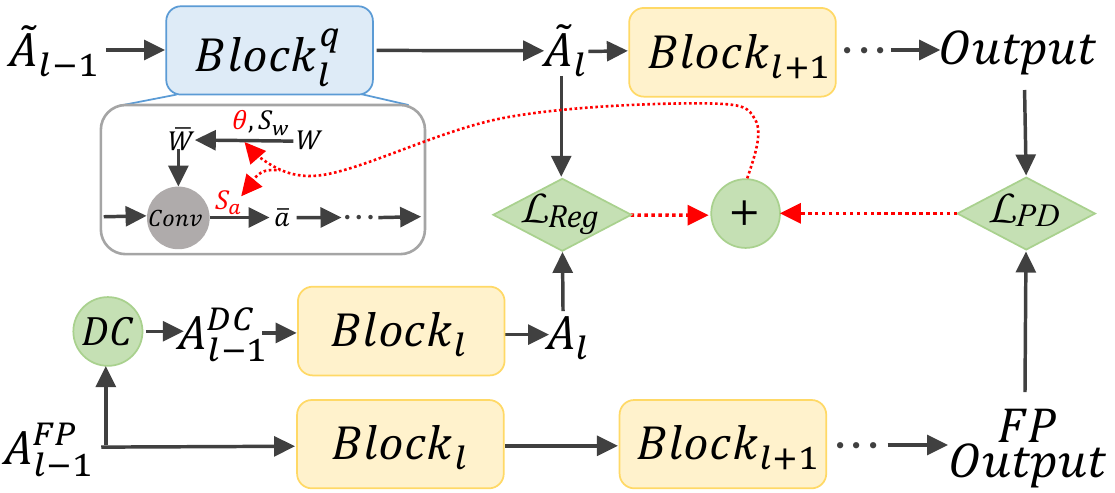}
   \caption{An overview of the PD-Quant. The blue and yellow rectangles indicate the quantized and FP layer, respectively. The green diamond is marked as the loss function. The green circle with DC indicates the Distribution Correction. FP output is the prediction of the whole FP network.}
   \vspace{-0.2cm}
   \label{fig:3.2_fig}
\end{figure}

To address the above overfitting problem, we introduce an explicit regularization~\cite{2017theory, hernandez2018data, qdrop} by adding a constraint to the optimization problem. 
Our regularization approach constrains the difference between the FP and quantized activation of the internal block.
On the one hand, the regularization can encourage the quantized model to learn the robustness from the FP model.
On the other hand, regularization is beneficial to minimize the perturbation caused by quantization, which is verified to be effective in BRECQ~\cite{brecq}.
Specifically, PD-Quant adopts PD loss to guide the quantization parameter optimization and introduces regularization to alleviate overfitting.
In our implementation, we regard block as the smallest unit as in previous work~\cite{brecq, qdrop, Mr.BiQ, yao2022rapq} and quantize neural networks from shallow to deep.
In more detail, when quantizing the $l_{th}$ block, our optimization objective is as follows:
\begin{align}\label{eq:recon_loss}
\begin{split}
 \arg\min\limits_{\theta, S_a} \ &\mathcal{L}_{PD}(O_{fp},\ f_{l+1}(\tilde{A}_{l}) + \lambda_r\mathcal{L}_{reg}(A_{l},\tilde{A}_{l})),\\
 &\tilde{A}_{l}={B}^{q}_{l}(\tilde{A}_{l-1};\ \theta, {S}_{a}),
\end{split}
\end{align}
where $\theta$ is described in equation \cref{eq:ada_quant}, and $S_a$ is the activation quantization scaling factor. 
$\lambda_r$ is a hyper-parameter to control the degree of regularization.
$\mathcal{L}_{reg}$ is the regularization loss to alleviate the overfitting problem.
Here we use the MSE distance between each block's output feature maps before and after quantization as $\mathcal{L}_{reg}=\| A_l - \tilde{A_l} \|_2^2$.
${B}^{q}_{l}$ denotes the block being quantized with input $\tilde{A}_{l-1}$. 
Like \cref{subsec:tradi_ptq}, we also make the quantization for all previous layers of the input $\tilde{A}_{l-1}$ has been done. 
$f_{l+1}(\cdot)$ is the part of the FP network mapping the output of quantizing block $\tilde{A}_{l}$ to the final prediction. 
$O_{fp}$ is the FP prediction as the target in PD loss. \cref{fig:3.2_fig} shows the overview of our PD-Quant.

By introducing regularization, the performance of the quantized model has been dramatically improved, shown in \cref{tab:3.2_tab} as PD+Reg.
The overfitting problem has been effectively alleviated, and the gap in performance between the calibration set and the validation set has narrowed a lot.
Random drop is also a regularization method, a supplement to our method for alleviating overfitting.
We introduce activation drop in the feature map as QDROP~\cite{qdrop}, which can further improve the performance of the quantized model.
The introduction of drop does not conflict with our method, and we only introduce it when computing the regularization loss.

\subsection{Distribution Correction for Regularization}
\label{subsec:dist_corr}
In this section, we will introduce a novel method to improve the generalizability of the quantized model further. 
Since only limited unlabeled images are accessible in PTQ, quantization parameters are determined only by activating these few samples.
However, the feature distribution of such small data is difficult to reflect the feature distribution of the whole training set.
As described in \cref{subsec:advanced_ptq}, our regularization loss computes the distance between the quantized block activation and the FP block activation.

\begin{figure}[tbp]
    \centering
    \subcaptionbox{ \label{without}}{\includegraphics[width=.48\linewidth]{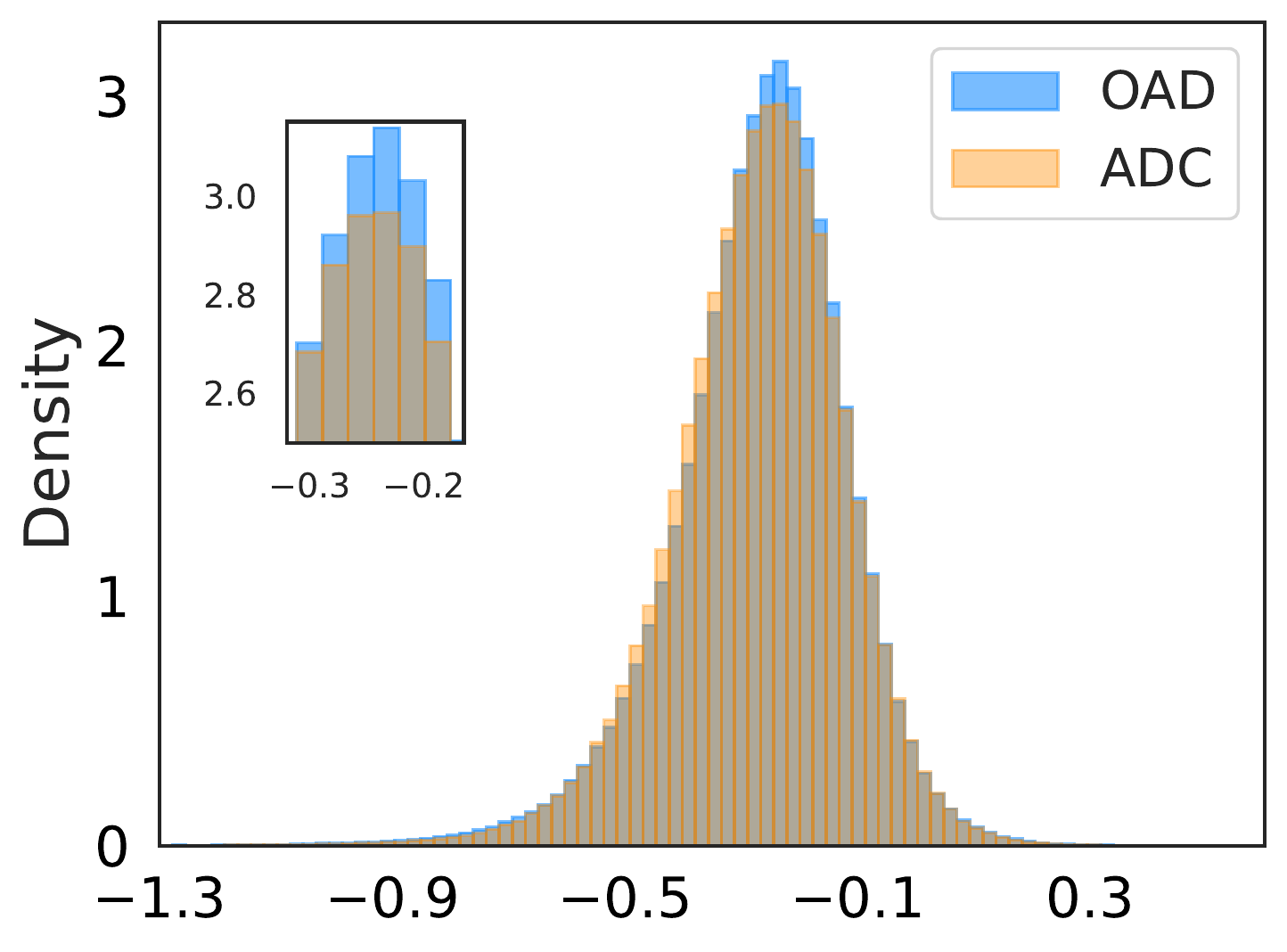}}
    \subcaptionbox{ \label{with1024}}{\includegraphics[width=.48\linewidth]{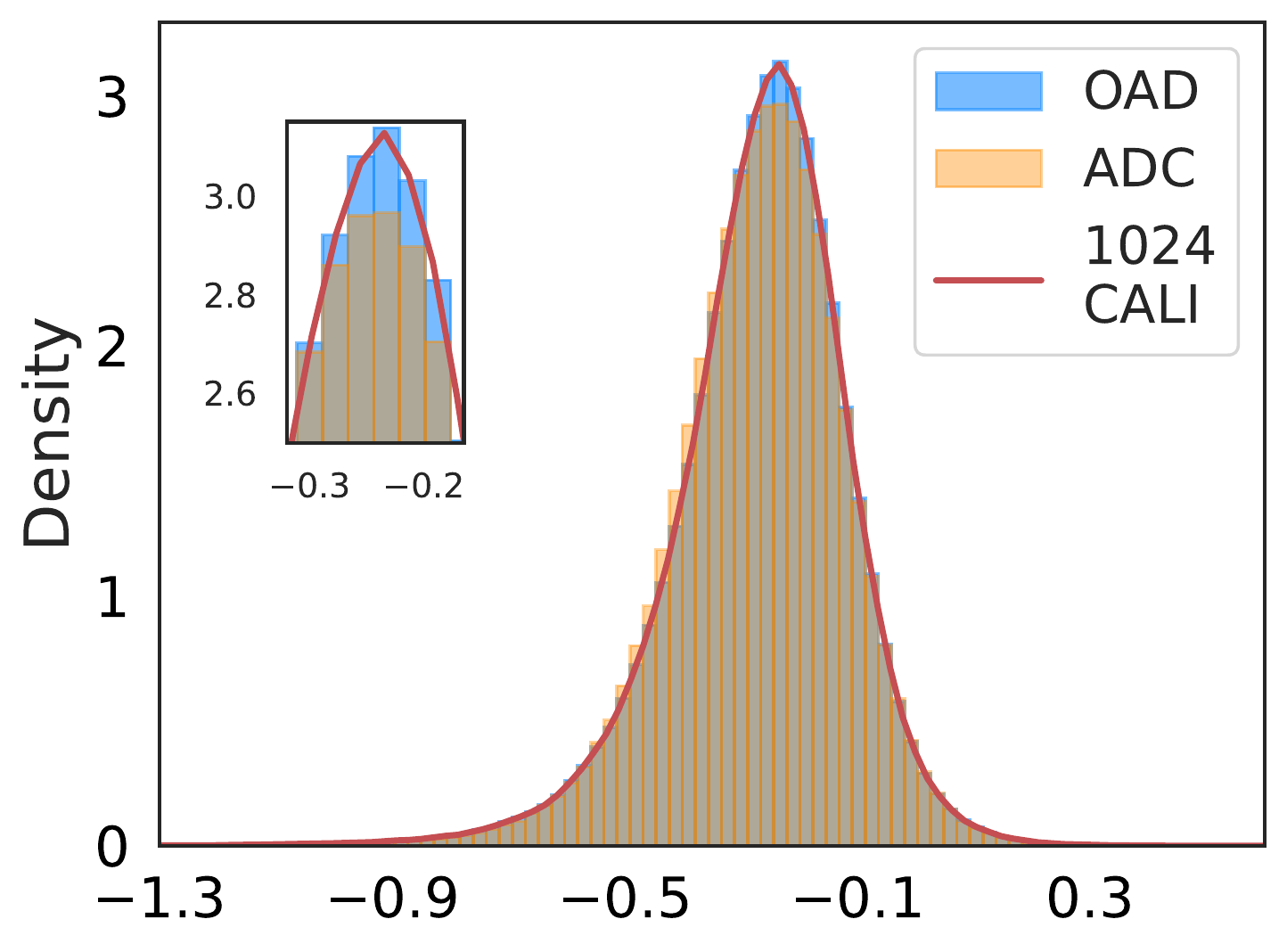}}\\
    \subcaptionbox{ \label{with10240}}{\includegraphics[width=.48\linewidth]{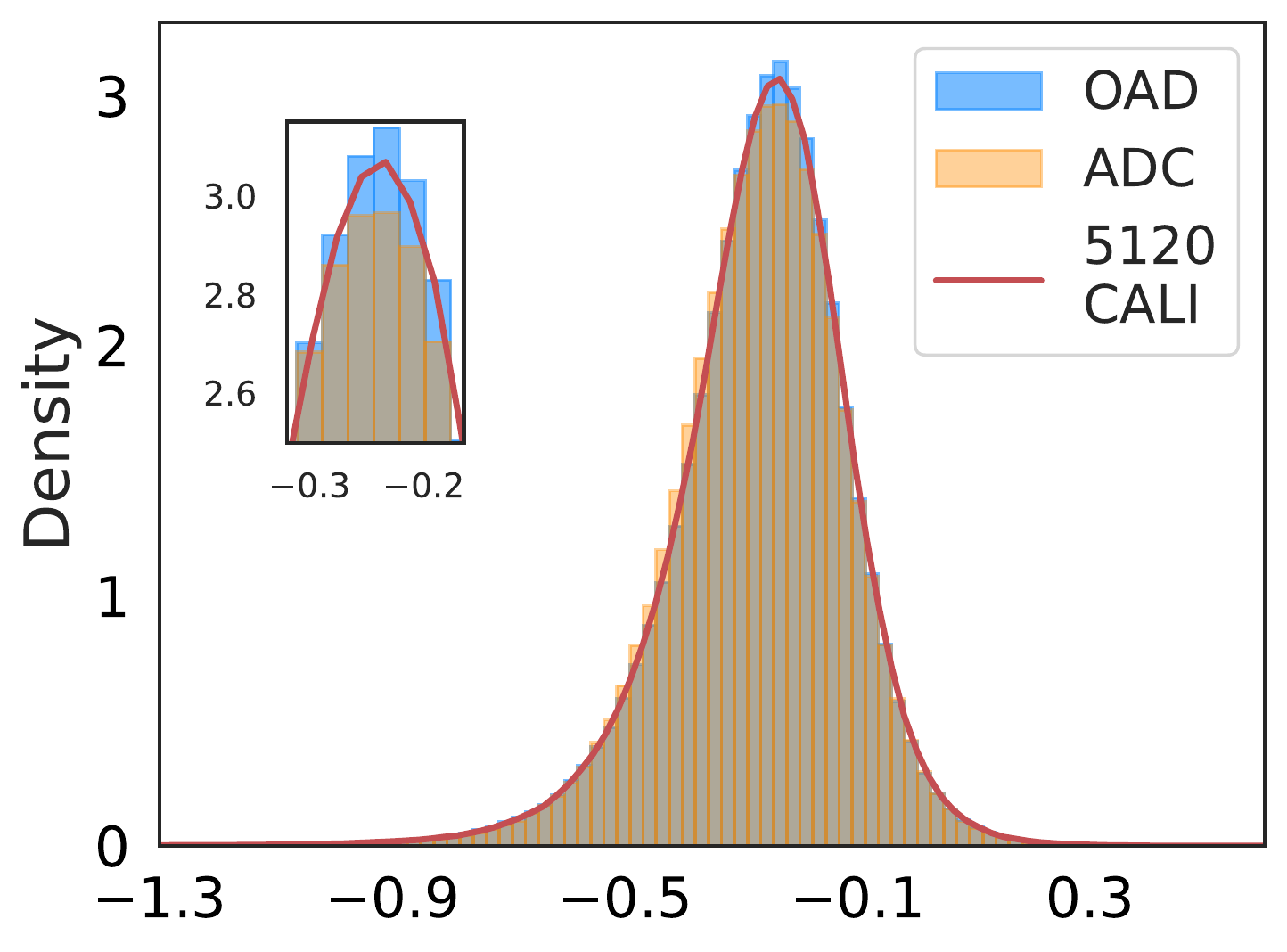}}
    \subcaptionbox{ \label{with51200}}{\includegraphics[width=.48\linewidth]{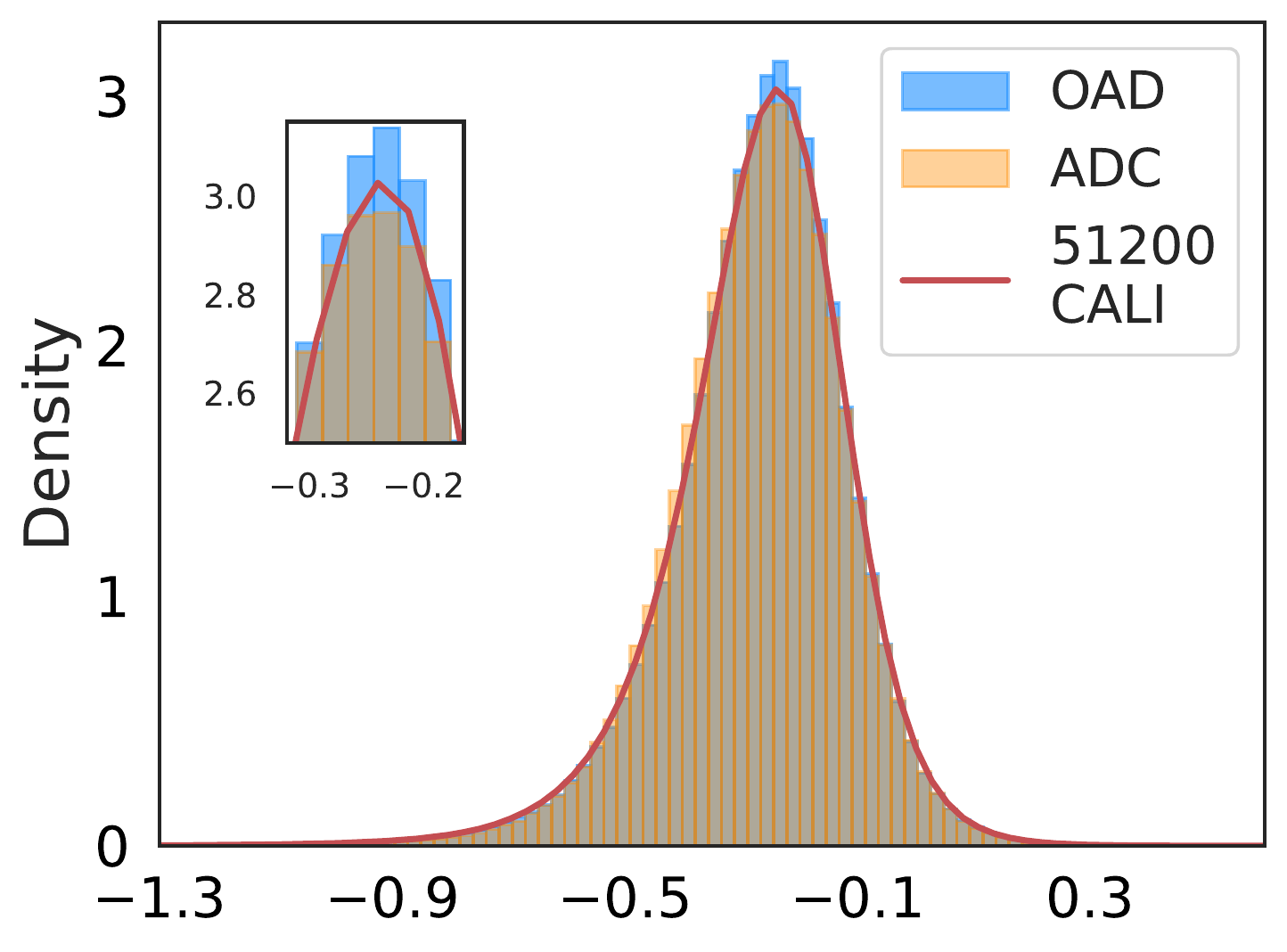}}
\caption{We plot the Distribution Correction for post-conv activation on the last block of ResNet18. OAD means the original activation distribution, and ADC denotes the original activation distribution (OAD) rectified by DC. Two histograms in blue and orange represent their distribution. The red line is the kernel density estimate (KDE) curve for different numbers of calibration(CALI) from the ImageNet training set.}
\vspace{-0.2cm}
\label{fig:hist}
\end{figure}

We propose a novel method to adjust the feature from the FP block itself, 
\ie, rectifying the feature by the statistics stored in the pre-trained FP model's Batch Normalization (BN) layer.
Specifically, we correct the activation distribution of each FP block's input by achieving an approximate mean and variance between the activation and BN statistics.
Since these BN statistics come from the whole training set, the corrected distribution of activations will be closer to the distribution of the entire training set.
Through learning the corrected FP feature distribution, the quantized model by DC has better generalizability.

\input{tables/performance_comparison}

Assuming there are $n$ Batch Normalization (BN) layers in the $l_{th}$ block, we can compute the mean and variance of their input, denoting $\{\hat\mu_{(i,l)}$, $\hat\sigma_{(i,l)} | i = 0,1...n\}$.
And the original mean ${\mu}_{(i,l)}$ and variance ${\sigma}_{(i,l)}$ are recorded in BN layers.
We modify the input of the block $A^{FP}_{l-1}$ to $A^{DC}_{l-1}$ using the following optimization:
\begin{align}\label{eq:bn_loss}
  \arg\min\limits_{A^{DC}_{l-1}}\ \lambda_c\sum_{i=1}^n&(\parallel \hat{\mu}_{(i,l)}-\mu_{(i,l)}\parallel_2^2+\parallel \hat{\sigma}_{(i,l)}-\sigma_{(i,l)}\parallel_2^2)\\
  \vspace{-1cm}
  &+ \parallel A^{DC}_{l-1}-A^{FP}_{l-1}\parallel_2^2, \nonumber
\end{align}
where $\lambda_c$ is a hyper-parameter. 
We fine-tuning the input $A^{DC}_{l-1}$ so that $\hat{\mu}_{(i,l)}$ and $\hat{\sigma}_{(i,l)}$ are closer to ${\mu}_{(i,l)}$ and ${\sigma}_{(i,l)}$.
The second term in \cref{eq:bn_loss} is to make the modified $A^{DC}_{l-1}$ not deviate too much from $A^{FP}_{l-1}$. 
As shown in \cref{fig:3.2_fig}, we inference with $A^{DC}_{l-1}$ to get $A_l$ and calculate $\mathcal{L}_{reg}$.

In \cref{fig:hist}, we visualize the effect of our proposed DC.
For all four subplots, the blue column (OAD) means an original activation distribution of 1024 calibration samples, and the orange column (ADC) denotes the OAD rectified by DC.
\cref{without} is the histogram of OAD and ADC. 
The red line in \cref{with1024} is the kernel density estimate (KDE) curve for 1024 calibration samples.
Next, we increase the number of calibration samples.
\cref{with10240} and \cref{with51200} show the KDE curve for 5120 and 51200 calibration samples from the ImageNet training set.
From \cref{fig:hist}, we notice that as the number of calibration samples increases, the KDE curve of calibration samples is closer to the ADC.
The illustration shows that the distribution of ADC, produced by the DC from OAD, can reflect the distribution of more samples in the training set and improve the generalizability of the quantized model.

%% file: tables/3.1_tab.tex
\begin{table}[b]
    \centering
     \resizebox{\columnwidth}{!}
     {
    \begin{tabular}{ccccccc}
    \toprule
    \textbf{Model} & \multicolumn{2}{c}{\textbf{ResNet-18}}&\multicolumn{2}{c}{\textbf{ResNet-50}} &\multicolumn{2}{c}{\textbf{RegNet-600M}}\\
    \cmidrule(lr){1-1}\cmidrule(lr){2-3}\cmidrule(lr){4-5}\cmidrule(lr){6-7}
    \textbf{Bits} & W8A2 & W4A2 & W8A2 & W4A2& W8A2 & W4A2 \\
    \midrule
Min-Max$^l$   &    -   &    -      &    -   &    -  &    -   &    -   \\
Cosine$^l$  &    11.09   &    4.15   &    2.19   &    1.14   &    0.96   &    0.65   \\
MSE$^l$   &    23.15   &   10.31   &    9.23   &    4.85  &    3.71   &    1.88   \\
\textbf{PD$^g$}   &    \textbf{28.41}   &    \textbf{12.27}   &    \textbf{11.31}   &    \textbf{6.01}&    \textbf{7.47}   &    \textbf{3.17}   \\

      \bottomrule
    \end{tabular} 
     } 
    \vspace{0.2mm}
    \caption{Metric test (top-1 accuracy($\%$) on validation set for activation scaling factors. The weight quantization for each bit setting in the table is the same. $^l$ represents calculating metric with only local information and $^g$ means metric with global information.}
    \vspace{0.2mm}
    \label{tab:3.1_tab}
\end{table}

%% file: tables/3.2_tab.tex
\begin{table}[b]

    \centering
    \resizebox{0.85\columnwidth}{!}
    { 
    
    \begin{tabular}{cccc}
    
    \toprule
   \textbf{Method} & \textbf{Bits (W/A)} &\textbf{Acc(val)} & \textbf{Acc(cali)}\\
    \midrule
    FP & 32/32 & 71.01 & 70.90\\
    \midrule

PD-only	& \multirow{3}{*}{2/2}    
	&		1.07	&	70.51	\\
\textbf{PD+Reg}&		&	\textbf{49.16}	&	71.09	\\
\textbf{PD+Reg+Drop}&		&	\textbf{52.74}	&	68.26		\\

\midrule
PD-only	&	\multirow{3}{*}{4/2}   
	&	51.32	&	70.41	\\
\textbf{PD+Reg}&		&	\textbf{56.20}	&	70.41	\\
\textbf{PD+Reg+Drop}	& &	\textbf{58.17}	&	68.36	\\

      \bottomrule
    \end{tabular} 
    }
    \vspace{0.2mm}
    \caption{PTQ accuracy on ImageNet at ResNet18 with 1024 calibration images. Reg means regularization. Acc(val)/Acc(cali) denotes the top-1 accuracy ($\%$) on validation/calibration set. }
    \vspace{0.2mm}
    
    \label{tab:3.2_tab}
\end{table}

%% file: tables/performance_comparison.tex
\begin{table*}[tb]
\centering
\footnotesize
\resizebox{0.99\textwidth}{!}
    {
    \begin{tabular}{cccccccc}
    \toprule
    \textbf{Methods} & \textbf{Bits (W/A)} & \textbf{ResNet-18} & \textbf{ResNet-50} & \textbf{MobileNetV2} & \textbf{RegNetX-600MF} & \textbf{RegNetX-3.2GF} & \textbf{MNasx2} \\
    \midrule
    Full Prec. & 32/32 & 71.01 & 76.63 & 72.62 & 73.52 & 78.46& 76.52\\
    \midrule
    ACIQ-Mix~\cite{aciq} & \multirow{6}{*}{4/4} & 67.00 & 73.80 & - & - & - & -\\
    LAPQ~\cite{lapq} &  & 60.30 & 70.00 & 49.70 & 57.71 & 55.89 & 65.32\\
    Bit-Split~\cite{wang2020towards} &  & 67.56 & 73.71 & - & - & - & -\\
    AdaRound~\cite{adaround} & & 67.96 & 73.88 & 61.52 & 68.20 & 73.85 & 68.86\\
    QDrop~\cite{qdrop}* &  & 69.17 & 75.15 & 68.07 & 70.91 & 76.40 & 72.81 \\
    \textbf{PD-Quant} &                     & \textbf{69.23}±0.06 & \textbf{75.16}±0.07 & \textbf{68.19}±0.12 & \textbf{70.95}±0.12 & \textbf{76.65}±0.09 & \textbf{73.26}±0.09 \\
    \midrule
    LAPQ & \multirow{4}{*}{2/4} & 0.18 & 0.14 & 0.13 & 0.17 & 0.12 & 0.18\\
    Adaround &  & 0.11 & 0.12 & 0.15 & - & - & -\\
    QDrop* &  & 64.57 & 70.09 & 53.37 & 63.18 & 71.96 & 63.23 \\
    \textbf{PD-Quant} &                     & \textbf{65.17}±0.08 & \textbf{70.77}±0.15 & \textbf{55.17}±0.28 & \textbf{63.89}±0.13 & \textbf{72.38}±0.11 & \textbf{63.40}±0.21 \\
    \midrule
    QDrop* & \multirow{2}{*}{4/2} & 57.56 & 63.26 & 17.30 & 49.73 & 62.00 & 34.12 \\
    \textbf{PD-Quant} &                     & \textbf{58.59}±0.15 & \textbf{64.18}±0.14 & \textbf{20.10}±0.37 & \textbf{51.09}±0.15 & \textbf{62.79}±0.13 & \textbf{39.13}±0.51 \\
    \midrule
    QDrop* & \multirow{2}{*}{2/2} & 51.42 & 55.45 & 10.28 & 39.01 & 54.38 & 23.59 \\
    \textbf{PD-Quant} &                     & \textbf{53.14}±0.14 & \textbf{57.16}±0.15 & \textbf{13.76}±0.40 & \textbf{40.67}±0.26 & \textbf{55.06}±0.23 & \textbf{27.58}±0.60 \\
    \bottomrule
    \end{tabular}}
    \vspace{0.2mm}
        \caption{Comparison on PD-Quant with various post-training quantization algorithms. 
        * denotes our implementation using open-source codes. PD-Quant is our proposed method.
        Other results listed are all from ~\cite{qdrop}.
        We gain the results of 10 runs using randomly sampled calibration sets.
        The results in the table include the mean and standard deviation.}
        \vspace{-0.2cm}
    \label{tab:performance_comparision}
\end{table*}

%% file: sections/experiment.tex
\section{Experimental Results}
\label{sec:results}

\subsection{Experimental Environments}
\label{subsec:environments}
We quantize various CNN architectures to evaluate our proposed method, including ResNet~\cite{resnet}, MobileNetV2~\cite{mobilenetv2}, RegNet~\cite{regnet}, and MNasNet~\cite{tan2019mnasnet}. 
The FP pre-trained models for all our implementation in the experiments are from~\cite{brecq}.
Our method evaluates on ImageNet dataset~\cite{russakovsky2015imagenet} with a batch size of 32.
We randomly sample 1024 images from the ImageNet training dataset as the calibration set. 
In addition, we also set the first and the last layer quantization to 8-bit for all PTQ experiments unless otherwise specified.
We keep the same quantization settings and hyper-parameters in our implementation as QDrop~\cite{qdrop}.
The learning rate for the activation quantization scaling factor is 4e-5, and for weight quantization rounding, the learning rate is 3e-3.
The DC is with a learning rate of 1e-3.
The choice of hyper-parameter $\lambda_r$ and $\lambda_c$ in \cref{eq:recon_loss} and \cref{eq:bn_loss} will be discussed later in \cref{subsec:sensitivity}, respectively.
Quantization parameters are fine-tuning with 20000 iterations.
Our code is based on Pytorch~\cite{paszke2019pytorch} and we evaluate all our experiments on a single Nvidia RTX A6000. 

\subsection{Performance Comparison}
\label{subsec:comparison}
We comprehensively compare our PD-Quant with multiple PTQ algorithms in many bit-settings and find that PD-Quant achieves performance improvement, especially in extremely low-bit.
Considering that only optimizing the activation scaling factors can not achieve usable results at low bits, all our experiments below optimize both rounding values and activation scaling factors unless otherwise specified.
We choose the best-performing method QDrop~\cite{qdrop} as our baseline.
The results of QDrop in open-source code are higher than those in their paper.
We introduced a drop~\cite{qdrop} in the regularization process as described in \cref{subsec:advanced_ptq}.
All our experiments with regularization include it unless otherwise stated.

In summarizing the results in~\cref{tab:performance_comparision}, it can be observed that PD-Quant achieves significant improvements compared with those strong baselines of PTQ.
When quantizing the network to W4A4, experiments show that PD-Quant slightly improves QDrop.
However, the benefits of PD-Quant become more apparent as the bit-width decreases.
At the W2A4 bit setting, the performance of PD-Quant is better than the baseline in all network architectures. 
With W4A2 quantization, PD-Quant can improve the accuracy of MobileNetV2 by 2.8\% and RegNet-600MF by 1.4\%.
For more challenging cases W2A2, the performance of PD-Quant surpasses QDrop in all networks.
According to the table, PD-Quant reaches 27.58\% in MNasNet, while QDrop only gets 23.59\%.
Since we use the same quantization setting with QDrop, the results indicate that the optimization strategy with our PD-Quant plays a critical role in extremely low-bit. 
Once PD-Quant finishes optimizing the quantization parameters, no additional computation is required for inference.

As in previous PTQ work, we also keep the first and last layer 8-bit.
The results in \cref{tab:performance_comparision} are all done with the 8-bit setting.
Nevertheless, to be noted, some work~\cite{brecq, Mr.BiQ, yao2022rapq} adds an extra first layer's output 8-bit, which will perform better than our 8-bit setting.
Therefore, we conduct further experiments to ensure the effectiveness of our method compared with these methods.
Specifically, we use the same 8-bit settings as these methods for experiments, and the results are shown in \cref{tab:brecq_comparision}.

\input{tables/brecq_comparison}

\subsection{Ablation Study}
\label{subsec:ablation}
\cref{tab:ablation} shows the ablation study of our proposed method compared with the baseline (QDrop)~\cite{qdrop}.
We quantize ResNet-18 and MobileNetV2 as examples to analyze the effects of PD and DC on the overall method.
We evaluate our method on W2A2 and W4A2, which can reveal the impact of each component.
PD-only optimizes the quantization parameters (activation scaling factors and rounding values) by only PD loss.
However, there is a huge drop in results because of over-fitting, as analyzed in \cref{subsec:advanced_ptq}.
We introduce Regularization to compensate for the performance loss of PD-only and achieve higher accuracy than the baseline.

Since QDrop optimizes the quantization parameters by computing the difference between the activation of the quantized block and the FP block.
So our proposed DC can be well applied to this baseline method to alleviate the overfitting problem.
Seeing the effect of QDrop+DC, it can improve by 0.9\% at W2A2 campared with the baseline.
In our PD-Quant, the introduction of DC can also further improve the performance by around 0.7\% in MobileNetV2 at the W2A2 setting.

The impact of introducing prediction difference can be seen from PD+Reg.
PD loss with regularization can improve the accuracy of ResNet-18 by 1.68\% and MobileNetV2 by 3.21\% compared with the baseline at the W2A2 bit setting.
In W4A2 quantization, the performance of our method is also better than the baseline.

\input{tables/ablation_study}

\subsection{Hyperparameters Analysis}
\label{subsec:sensitivity}
As described in \cref{eq:recon_loss}, there is a hyper-parameter $\lambda_r$ in the optimization objective.
$\lambda_r$ limits the strength of the regularization constraint.
We analyze how it affects quantization performance on ResNet-18 and MobileNetV2 with W2A2 quantization.
According to \cref{tab:pd_ablation}, we found that different networks have different optimal $\lambda_v$, but our method is not very sensitive to it.
Results show that the proposed method can achieve a steady accuracy without tedious hyperparameter tuning.

\input{tables/pd_ablation}

\input{tables/dc_ablation}

Moreover, we also conduct experiments to analyze the hyperparameter in Distribution Correction. 
As illustrated in \cref{eq:bn_loss}, $\lambda_c$ controls the degree of correction.
We demonstrate how $\lambda_c$ affects model performance in \cref{tab:dc_ablation}, where $\lambda_c=0$ means no DC.

Although there is some effect when $\lambda_c$ changes, our DC method can also get a stable level of accuracy without hand-crafted hyper-parameter adjusting.

%% file: tables/brecq_comparison.tex
\begin{table}[htb]
\centering
\footnotesize
\resizebox{0.95\columnwidth}{!}
    {
    \begin{tabular}{cccc}
    \toprule
    \textbf{Methods} & \textbf{Bits (W/A)} & \textbf{ResNet-18} & \textbf{MobileNetV2} \\
    \midrule
    Full Prec. & 32/32 & 71.01 & 72.62\\
    \midrule
    BRECQ~\cite{brecq}  &  \multirow{3}{*}{4/4}  &   69.60 & 66.57\\
    RAPQ~\cite{yao2022rapq} &    &   69.28  &   64.48 \\
    \textbf{PD-Quant}&    &   \textbf{69.72}    &   \textbf{68.76} \\
    \midrule
    BRECQ~\cite{brecq}  &   \multirow{3}{*}{2/4} &   64.80   & 53.34  \\
    RAPQ~\cite{yao2022rapq} &    &   65.32   &   48.12 \\ 
    \textbf{PD-Quant}&    &   \textbf{65.56}  &   \textbf{55.32} \\
    \bottomrule
    \end{tabular}}
    \vspace{-1mm}
        \caption{Comparison on PD-Quant with different post-training quantization algorithms in another 8-bit setting. }
        \vspace{-0.1cm}
    \label{tab:brecq_comparision}
\end{table}

%% file: tables/ablation_study.tex
\begin{table}[thb]
    \centering
     \resizebox{0.85\columnwidth}{!}
     {
    \begin{tabular}{ccccc}
    \toprule
    \textbf{Model} & \multicolumn{2}{c}{\textbf{ResNet-18}}&\multicolumn{2}{c}{\textbf{MobileNetV2}}\\
    \cmidrule(lr){1-1}\cmidrule(lr){2-3}\cmidrule(lr){4-5}
    \textbf{Bits} & W2A2 & W4A2 & W2A2& W4A2\\
    \midrule

QDrop   &    51.42   &    57.56   &    10.28     &    17.30   \\
PD-only &    1.07   &    51.32   &    7.01    &    13.59   \\
PD+Reg   &    52.74   &   58.17   &    13.49     &    20.05   \\
QDrop+DC  &    52.32   &   57.77   &    10.38     &    17.58   \\
\textbf{PD-Quant}  &    \textbf{53.08}   &    \textbf{58.65}   &    \textbf{14.17}   &    \textbf{20.40}   \\
      \bottomrule
    \end{tabular} 
     } 
    \vspace{0.2mm}
    \caption{Ablation study (top-1 accuracy($\%$)) on validation set for our proposed method. QDrop is the baseline method. PD-only means optimizing quantization parameters by only PD loss. Reg means regularization. PD-Quant is our proposed method, including PD, Reg, and DC for optimizing both activation scaling factors and rounding values. }
    \vspace{-1mm}
    \label{tab:ablation}
\end{table}

%% file: tables/pd_ablation.tex
\begin{table}[htb!]
    \centering

     \resizebox{0.9\columnwidth}{!}
     {
    \begin{tabular}{cccccc}
    \toprule
    \multirow{2}{*}{\textbf{Model}}  & \multicolumn{5}{c}{$\pmb{\lambda_r}$}\\
    \cmidrule(lr){2-6}

   & 0.05 & 0.1 & 0.2 & 0.5 & 1  \\
    \midrule
    
ResNet-18 	& {52.43} & {52.60} & {52.74} & {52.58} & {52.55} \\
MobileNetV2   & {13.48} & {13.49} & {13.03} & {13.37} & {12.38} 	\\

      \bottomrule
    \end{tabular} 
     } 
    \vspace{-1mm}
    \caption{Hyperparameter analysis for $\lambda_r$ at W2A2 setting.}
    \vspace{-1mm}
    \label{tab:pd_ablation}
\end{table}

%% file: tables/dc_ablation.tex
\begin{table}[htb!]
    \centering

     \resizebox{0.9\columnwidth}{!}
     {
    \begin{tabular}{cccccc}
    \toprule
    \multirow{2}{*}{\textbf{Model}}  & \multicolumn{5}{c}{$\pmb{\lambda_c}$}\\
    \cmidrule(lr){2-6}
    & 0 & 0.001 & 0.005 & 0.01 & 0.02  \\
    \midrule

ResNet-18    	& {52.74} & {53.04} & {52.87} & {52.89} &	{53.08} \\
MobileNetV2    & {13.49} & {13.77} & {14.17} & {13.24} & {13.09} \\

      \bottomrule
    \end{tabular} 
     } 
    \vspace{-1mm}
    \caption{Hyperparameter analysis for $\lambda_c$ at W2A2 setting.}
    \label{tab:dc_ablation}
\end{table}

%% file: sections/discussion.tex
\section{Discussion}
\label{sec:discussion}
In this section, we will first discuss why we do not determine weight quantization scaling factors by PD loss.
Then the reasons for choosing KL to calculate the prediction difference will be explained.
Finally, we will analyze the limitations of PD-Quant.

\subsection{Weight Quantization Scaling Factor With PD Loss}
\label{subsec:weiht_quantize}
We didn't use PD loss to find weight quantization scaling factors ($S_w$) in \cref{subsec:tradi_ptq} because weight quantization is different from activation quantization.
Current methods mostly quantize weights per channel, resulting in a much larger search space than activation quantization.
Besides, our experiments in \cref{tab:weight_discuss} show that the weight scaling factors are not sensitive to the introduction of PD information.

\input{tables/weight_discuss}

We evaluate how PD loss affects activation scaling factors and weight scaling factors in \cref{tab:weight_discuss}.
To avoid the influence of other quantization parameters, we optimize only scaling factors in the experiments.
As can be seen, all existing methods fall crash, including PD loss, when we quantize the weights with low bits.
To better understand the impact of our method on weight quantization, we conducted an experiment on the first layer only, referred to as "First Layer" for the rest of the paper. 
Unlike the previous experiments, the First Layer did not set the first and last layer to 8 bit, and we set per-tensor weight quantization in the First Layer of W2A32.
Our method can effectively search for the optimal scaling factor during activation quantization at low bits, but not for weight quantization.
We think this is because the parameter space of weights is deterministic, while the parameter space of activations varies with different input image.

\subsection{Determination of PD loss}
\label{subsec:pd_determination}
In \cref{tab:pd_determination}, we have compared different global metrics for searching the activation quantization scaling factors. 
MSE$^g$ and Cosine$^g$ measure the difference between the quantized prediction and FP prediction by MSE distance and cosine distance. 
KL$^g$ consider the KL divergence~\cite{hinton2015distilling} of the two models' prediction.
As can be seen in \cref{tab:pd_determination}, KL performs best among all metrics.
We believe this is because our quantization process can be seen as a knowledge distillation~\cite{hinton2015distilling, gou2021knowledge}, where the quantized model is the student and the FP model is the teacher.
\input{tables/pd_determination}

\subsection{Limitations}
\label{subsec:limitations}
PD-Qunat requires additional computation and fine-tuning compared to the baseline method QDROP, resulting in increased time cost.
The time cost of quantization is shown in \cref{tab:cost}.
While training time consumption is a limitation of PD-Quant, our method's additional time cost is acceptable when compared to quantization-aware training (QAT) methods. 
For example, LSQ~\cite{lsq}, a QAT method, takes 120 hours to train for 90 epochs at ResNet-18, while PD-Quant only needs 1 hour.

\input{tables/cost}

%% file: tables/weight_discuss.tex
\begin{table}[thb]
    \centering
     \resizebox{0.9\columnwidth}{!}
     {
    \begin{tabular}{ccccc}
    \toprule
    \multirow{2}{*}{\textbf{Metric}}    &\multicolumn{2}{c}{\textbf{W2A32}}  &\multicolumn{2}{c}{\textbf{W32A2}}\\
    \cmidrule(lr){2-3}\cmidrule(lr){4-5}
     & First layer & All layers & First layer & All layers\\
    \midrule    
        MSE$^l$ &  \textbf{11.09}  &   0.09 &    56.10&    23.38\\
        PD$^g$  &   10.98   &0.10&    \textbf{56.83}&    \textbf{28.66}      \\
      \bottomrule
    \end{tabular} 
     } 
    \vspace{1mm}
    \caption{We quantize only weights/activations for ResNet-18 to investigate how PD loss affects weight/activation quantization scaling factors. The results listed include only quantizing the first layer of the model and quantizing all layers.}
    \vspace{-0.2cm}
    \label{tab:weight_discuss}
\end{table}

%% file: tables/pd_determination.tex
\begin{table}[bht]
    \centering
     \resizebox{0.8\columnwidth}{!}
     {
    \begin{tabular}{ccccc}
    \toprule
    \textbf{Model} & \multicolumn{2}{c}{\textbf{ResNet-18}} &\multicolumn{2}{c}{\textbf{RegNet-600M}}\\
    \cmidrule(lr){1-1}\cmidrule(lr){2-3}\cmidrule(lr){4-5}
    \textbf{Metric} & W8A2 & W4A2& W8A2 & W4A2 \\
    \midrule
MSE$^g$     &    27.06   &    11.13   &  3.70    &   1.64   \\
Cosine$^g$  &    27.28   &    7.12     &    5.93  &    3.06  \\
\textbf{KL$^g$}   &    \textbf{28.41}   &    \textbf{12.27}   &      \textbf{7.47}   &    \textbf{3.17}   \\

      \bottomrule
    \end{tabular} 
     } 
    \vspace{1mm}
    \caption{Metric test (top-1 accuracy($\%$) on validation set for activation scaling factors. $^g$ means calculating the difference between predictions.}
    \vspace{-0.1cm}
    \label{tab:pd_determination}
\end{table}

%% file: tables/cost.tex
\begin{table}[htb]
    \centering
     \resizebox{\columnwidth}{!}
     {
    \begin{tabular}{cccc}
    \toprule
    \textbf{Method} & \textbf{ResNet-18} & \textbf{MobileNetV2} & \textbf{RegNetX-600MF}\\
    \midrule
QDrop   &    0.43h   &    0.93h   &    0.89h   \\
PD   &    0.91h   &   2.26h   &    2.37h   \\
PD+DC & 1.11h & 2.68h &  2.75h \\
      \bottomrule
    \end{tabular} 
     } 
    \vspace{-0.2cm}  
    \caption{Time cost comparison. (one Nvidia RTX A6000)}
    \vspace{-0.4cm}
    \label{tab:cost}
\end{table}

%% file: sections/conclusion.tex
\section{Conclusion}
\label{conclusion}
In this work, we first observed that it was very beneficial to introduce PD information when optimizing the activation scaling factors at low-bit.
Based on this observation, we proposed PD-Quant, an effective method for post-training quantization.
When optimizing activation scaling factors and rounding values, we discovered that our proposed method could also improve current methods.
In addition, we found that over-fitting is a factor that leads to performance degradation.
To respond to this problem, we propose a method to correct the calibration set distribution to improve model generalizability.
The experimental results show that PD-Quant is very effective at low bits.

%% file: sections/supplementary.tex
\begin{appendices}
  
\section{More Details of CNN models Implementations}
This section will add more experimental details for CNN models.
We apply different hyper-parameters $\lambda_r$ and $\lambda_c$ for different types of networks.
The regularization parameter $\lambda_r$ is set to 0.2 for ResNet-18 and ResNet-50~\cite{resnet} and 0.1 for other CNN architectures.
Moreover, we set the hyper-parameter $\lambda_c$ for DC to 0.005 for MobileNetV2~\cite{mobilenetv2}, 0.001 for MNasNet~\cite{tan2019mnasnet}, and 0.02 for other CNN architectures.

\section{Effects on different calibration data sizes}
We conduct experiments on 256, 1024, and 4096 calibration data sizes.
\cref{tab:calib_size_ablation} shows that PD is effective on calibration data of different sizes.
The effect of DC decreases as the size of calibration sets increases because the calibration set's distribution is getting closer to the training set.
\input{tables/calib_size_ablation}

\section{PD Loss on Transformer Models}
Besides CNN, we further extend the proposed method to Transformer models.
We evaluate our PD-Quant on both ViT~\cite{dosovitskiy2020image} and DeiT~\cite{touvron2021training} at different bit settings.

\subsection{Implementation Details}
We keep most parameter settings the same as in CNN, including the learning rate, iterations, and calibration data numbers.
However, we set the batch size to 16 and regularization parameters $\lambda_r$ to 0.1 for Transformer models.
We did not apply DC to the quantization of Transformer models because there are no batch normalization layers.

We quantize all the weights and inputs for the fully-connect layers, including the first projection layer and the last head layer.
The two input matrices for the matrix multiplications in the self-attention modules are also quantized.
Moreover, the inputs of the softmax layers and the normalization layers are not quantized, the same as in previous work~\cite{liu2021post, yuan2022ptq4vit}.

We still take QDrop as the baseline method and define the encoder in Transformer models as the block.
Our implementation for Transformer models is based on open-source code, and the pre-trained FP models are all from~\cite{rw2019timm}.

\input{tables/vit_comparison.tex}

\subsection{Performance Comparison}
We compare our proposed PD-Quant with QDrop~\cite{qdrop} and PTQ4ViT~\cite{yuan2022ptq4vit} for both ViT and DeiT.
PQT4ViT is a post-training quantization framework designed for Transformer model quantization.
Moreover, it shows the state-of-the-art results among all transformer quantization algorithms in W6A6.
We keep the same quantization environment and use the same pre-trained model for comparison.

As seen in \cref{vit_compare}, PD-Quant can improve the results of QDrop, similar to the effects in CNN models.
We implemented PTQ4ViT based on open-source code.

\section{Optimization of Activation Scaling Factors and Rounding values}
QAT method LSQ~\cite{lsq} first optimizes activation scaling factors ($S_a$) by final objective.
Since only limited unlabeled data is available in PTQ, we propose PD loss to optimize $S_a$.
When optimizing only $S_a$, the gradients are given by

\begin{equation}
    \FracPartial{\mathcal{L}_{PD}}{S_a} = 
    \left\{
    \begin{aligned}
    &\FracPartial{\mathcal{L}_{PD}}{\Tilde{x}}q_{max}& &\frac{x}{S_a} \geq q_{max}\\
    &\FracPartial{\mathcal{L}_{PD}}{\Tilde{x}}\left(\lfloor\frac{x}{S_a}\rceil-\frac{x}{S_a}\right)& &{q_{min} < \frac{x}{S_a} < q_{max}}\\
    &\FracPartial{\mathcal{L}_{PD}}{\Tilde{x}}q_{min}& &{\frac{x}{S_a} \leq q_{min}}
    \end{aligned}
    \right.,
    \label{eq:lsq}
\end{equation}
where STE~\cite{ste} calculates the gradients of the round function.

When optimizing rounding values ($\theta$), we follow~\cite{adaround} to adopt a sigmoid-like function $\sigma(\theta)$ deciding weight rounding up or down.
The minimization problem for $\theta$ convergence is given by
\begin{align}\label{eq:adaround}
 \arg\min\limits_\theta \sum(1-|2\sigma(\theta)-1|^\beta),
\end{align}
where $\sigma(\theta)=0$ means weight rounds down and $\sigma(\theta)=1$ means weight rounds up.

\end{appendices}

%% file: tables/calib_size_ablation.tex
\begin{table}[thb]
    \centering
    \vspace{-3mm}
     \resizebox{\columnwidth}{!}
     {
    \begin{tabular}{ccccccc}
    \toprule
    \textbf{Model} & \multicolumn{3}{c}{\textbf{ResNet-18}}&\multicolumn{3}{c}{\textbf{MobileNetV2}}\\
    \cmidrule(lr){1-1}\cmidrule(lr){2-4}\cmidrule(lr){5-7}
    \textbf{Size} & 256 & 1024 & 4096 & 256 & 1024 & 4096\\
    \midrule

QDrop   &    46.22   &    51.42   &    54.48    &    7.53 & 10.28 &  10.88\\
PD &    46.76   &    52.74   &    55.30    &    9.29 & 13.49 &  16.47\\
\textbf{PD+DC}  &    \textbf{47.28}   &    \textbf{53.08}   &    \textbf{55.33}   &    \textbf{9.48} &    \textbf{14.17} &    \textbf{16.55}\\
      \bottomrule
    \end{tabular} 
     }
    \vspace{-0.2cm}
    \caption{Effects on different calibration dataset sizes for PD-Quant. All the results in the table are quantized to W2A2.}
    \vspace{-0.4cm}
    \label{tab:calib_size_ablation}
\end{table}

%% file: tables/vit_comparison.tex
\begin{table}[tb]
\setlength\tabcolsep{6pt}
\centering

\resizebox{\linewidth}{!}{
\begin{tabular}{@{}cccc@{}}
\toprule
\textbf{Model}                & \textbf{Method}     & \textbf{Bits (W/A)}  & \textbf{Acc (\%)}  \\ 
\midrule
\multicolumn{1}{l}{}                & PTQ4ViT*~\cite{yuan2022ptq4vit}  & \multirow{3}{*}{W6A6}      & 70.72  \\
\multicolumn{1}{l}{}                     & QDrop*~\cite{qdrop}        &      & 70.25  \\
                    & \textbf{PD-Quant}        &  & \textbf{70.84}  \\
\cmidrule(l){2-4} 
 
               & PTQ4ViT*~\cite{yuan2022ptq4vit}  & \multirow{3}{*}{W4A6}      & 53.55  \\
ViT-S/16/224                      & QDrop*~\cite{qdrop}        &       & 67.57  \\
74.65                     & \textbf{PD-Quant}        &  & \textbf{68.64}  \\
\cmidrule(l){2-4} 
                & PTQ4ViT*~\cite{yuan2022ptq4vit}  & \multirow{3}{*}{W2A6}      & 0.31  \\
                     & QDrop*~\cite{qdrop}    &         & 45.16  \\
                     & \textbf{PD-Quant}    &      & \textbf{48.13}  \\
\midrule
\multicolumn{1}{l}{}                & PTQ4ViT*~\cite{yuan2022ptq4vit}  & \multirow{3}{*}{W6A6}     & 74.24  \\
\multicolumn{1}{l}{}                     & QDrop*~\cite{qdrop}        &       & 75.76  \\
                     & \textbf{PD-Quant}        &   & \textbf{75.82}  \\
\cmidrule(l){2-4} 
                & PTQ4ViT*~\cite{yuan2022ptq4vit}  & \multirow{3}{*}{W4A6}      & 52.97  \\
ViT-B/16/224                     & QDrop*~\cite{qdrop}        &       & 75.51  \\
78.01                     & \textbf{PD-Quant}        &   & \textbf{75.52}  \\
\cmidrule(l){2-4} 
                & PTQ4ViT*~\cite{yuan2022ptq4vit}  & \multirow{3}{*}{W2A6}      & 0.24  \\
                     & QDrop*~\cite{qdrop}        &       & 63.74  \\
                     & \textbf{PD-Quant}        &   & \textbf{64.51}  \\
\midrule
\multicolumn{1}{l}{} & PTQ4ViT*~\cite{yuan2022ptq4vit}  & \multirow{3}{*}{W6A6}      & 76.83  \\
\multicolumn{1}{l}{} & QDrop*~\cite{qdrop}        &       & 77.95  \\
         & \textbf{PD-Quant}    &       & \textbf{78.33}  \\ \cmidrule(l){2-4} 
                & PTQ4ViT*~\cite{yuan2022ptq4vit}  & \multirow{3}{*}{W4A6}      & 74.17  \\
DeiT-S/16/224                     & QDrop*~\cite{qdrop}        &       & 77.66  \\
79.71                     & \textbf{PD-Quant}        &   & \textbf{77.88}  \\
\cmidrule(l){2-4} 
                & PTQ4ViT*~\cite{yuan2022ptq4vit}  & \multirow{3}{*}{W2A6}      & 3.79  \\
                     & QDrop*~\cite{qdrop}        &       & 65.76  \\
                     & \textbf{PD-Quant}        &   & \textbf{67.53}\\
\bottomrule

\end{tabular}
}
\label{vit_compare}
\caption{Comparison on PD-Quant for Transformer models. 
* represents our implementation with open-source code.
ViT-S/16/224 denotes patch size is $16\times16$ the input resolution is $224\times224$.
All the results listed are the top-1 accuracy (\%).}
\vspace{-0.5cm}
\end{table}